\documentclass{article}
\usepackage[final]{corl_2017} 

\usepackage{graphicx} 
\usepackage{subfigure} 
\usepackage{wrapfig}

\usepackage{natbib}

\usepackage{algorithm}
\usepackage{algorithmic}

\newcommand{\rT}{{\mathrm{T}}}

\usepackage{amsmath} 
\usepackage{amssymb}  

\title{Aggressive Deep Driving: Model Predictive Control with a {CNN} Cost Model}

\author{
  Paul Drews\\
  School of ECE \\
  Georgia Institute of Technology \\
  \texttt{pdrews3@gatech.edu} \\
  \And
  Grady Williams \\
  College of Computing\\
  Georgia Institute of Technology \\
  \texttt{gradyrw@gatech.edu} \\
  \AND
  Brian Goldfain \\
  College of Computing\\
  Georgia Institute of Technology \\
  \texttt{bgoldfain3@gatech.edu} \\
  \And
  Evangelos A. Theodorou \\
  School of Aerospace Engineering\\
  Georgia Institute of Technology \\
  United States \\
  \texttt{evangelos.theodorou@gatech.edu} \\
  \And
  James M. Rehg \\
  College of Computing\\
  Georgia Institute of Technology \\
  \texttt{rehg@gatech.edu} \\
}

\begin{document}
\maketitle


\begin{abstract} 
We present a framework for vision-based model predictive control (MPC) for the task of aggressive, high-speed autonomous driving. Our approach uses deep convolutional neural networks to predict cost functions from input video which are directly suitable for online trajectory optimization with MPC. We demonstrate the method in a high speed autonomous driving scenario, where we use a single monocular camera and a deep convolutional neural network to predict a cost map of the track in front of the vehicle. Results are demonstrated on a 1:5 scale autonomous vehicle given the task of high speed, aggressive driving.
\end{abstract}





\section{Introduction}
A basic challenge in autonomous driving is to couple perception and control in order to achieve a desired vehicle trajectory in the face of uncertainty about the environment. Existing commercial solutions for driver assistance and vehicle autonomy utilize relatively simple models of vehicle dynamics, and emphasize the integration of multiple sensing modalities to characterize the vehicle's environment. Several examples of this approach can be found in the perception and control architectures utilized in the DARPA Grand Challenge Competitions \cite{junior2008,urmson2008boss,urmson2007tartan}. 

%
%


While many challenging problems remain in order to achieve safe and effective autonomous driving in urban environments, this paper is focused on the task of \emph{aggressive driving}, which requires a tight coupling between control and perception. We define aggressive driving as a vehicle operating close to the limits of handling, often with high sideslip angles, such as may be required for collision avoidance or racing. There has recently been some prior work on aggressive driving using a 1:5 scale vehicle~\cite{williams2017information,williams2016}. This work resulted in an open source vehicle platform which we have also adopted in this paper. A disadvantage of this prior work is its reliance on high-quality GPS and IMU for position estimation and localization, which limits the applicability of the method. In this paper, we present an approach to auonomous racing in which vehicle control is based on computer vision sensing, using only monocular camera images acquired from a dirt track in a rally car racing environment. We address the challenge of learning visual models which can be executed in real-time to support high-speed driving. We make the following contributions:
\begin{itemize}
	\item A novel deep learning approach for analyzing monocular video and generating real-time cost maps for model predictive control which can drive an autonomous vehicle at high speeds
  \item Analysis of the benefits of different representations for the traversability map, with the demonstration that direct prediction of an overhead (bird's eye view) cost map gives the best performance
  \item A method for automatic image annotation to support large-scale human-in-the-loop training of deep neural networks for autonomous driving
\end{itemize}

Our framework is able to take as input a single monocular camera image and output a costmap of the area in front of the vehicle.  This costmap image is fed directly into a model predictive control algorithm, with no pre-processing steps necessary.  Because the costmap learned by the neural network is independent of the control task being performed, we can use any driving data, including human data, as training data and still generalize to different tasks.  Additionally, because we learn an interpretable intermediate representation, it is much easier to diagnose failure cases.  For example, in Figure \ref{fig:velocities}, it is clear the network is confusing an upcoming right turn with a left turn.

\subsection{Related Work}

\begin{figure*}
\vskip 0.2in
\begin{center}
\includegraphics[width=0.4\columnwidth]{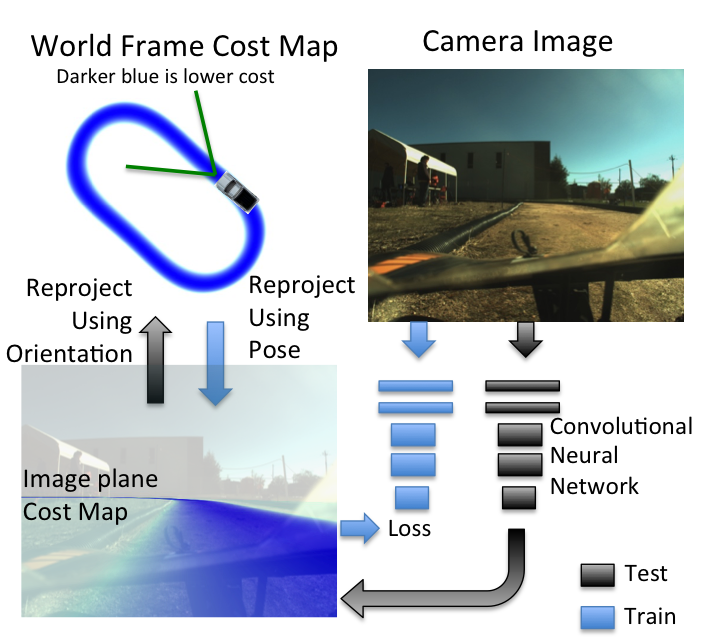}
\includegraphics[width=0.4\columnwidth]{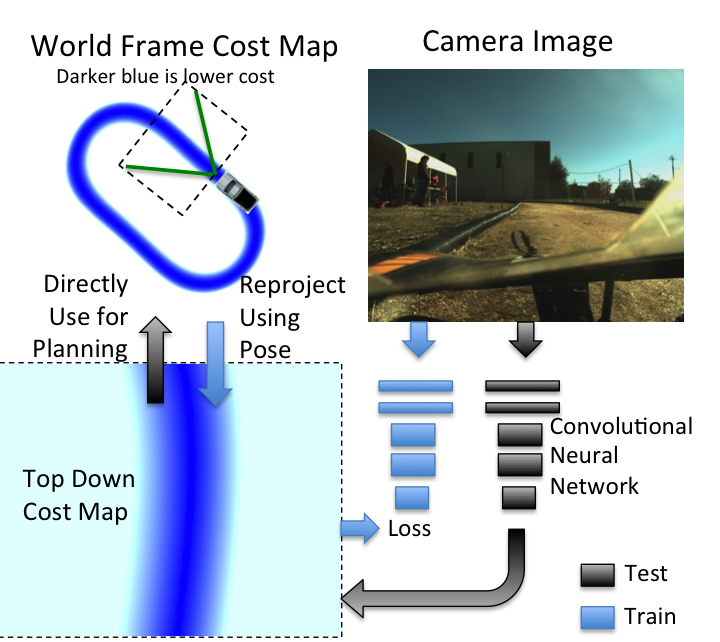}
\caption{Automatic image labeling for neural network training}
\label{fig:overview}
\end{center}
\vskip -0.2in
\end{figure*}

Several approaches have been taken to solve the problem of aggressive autonomous driving, and autonomous driving in general.  In \cite{funke2012limits}, an analytic approach is explored.  The performance limits of a vehicle are pushed using a simple model-based feedback controller and extensive pre-planning to follow a racing line around a track. More recently, \cite{Keivan2014} showed the benefits of model predictive control on a 1:10 scale vehicle following waypoints through a challenging obstacle course.  \cite{williams2016} also shows some of the benefits of model predictive control in an outdoor, dirt environment.

However, these approaches all rely on highly accurate position from an external source (either GPS or motion capture).  When approaching autonomous driving in natural environments, we would prefer being able to drive aggressively using only internal sensors, and ideally only low-cost sensors such as cameras and IMU.  There are several ways to approach this problem.

There are many SLAM approaches that use cameras \cite{engel2014lsd,mur2015orb}, LIDAR \cite{zhang2014loam}, or other sensor combinations \cite{newcombe2011kinectfusion} to provide accurate position.  These systems typically provide position relative to a generated map.  However, this approach can be very challenging when localizing in a map created in significantly different conditions \cite{beall2014appearance}.  Because a large map needs to be created, and position calculated relative to this map, these methods tend to be computationally expensive.  An alternative method to providing absolute position uses deep neural networks to directly regress a position estimate in an area previously visited \cite{kendall2015modelling}.  However, this method of localization is not yet sufficiently accurate to be directly used for control.  Our method replaces the need for any type of direct positioning or pre-computation of 

Instead of relying directly on accurate localization, one can instead rely on camera images to derive actions, bypassing the need for explicit position information.  In \cite{bojarski2016end}, a strong case is made for end to end learning, or behavior reflex control in the context of autonomous driving.  In this paper, the authors train a neural network to directly regress a steering control signal from images, given only training data of humans driving on many different types of roadways.  This work follows from seminal work by Dean Pomerleau in the Alvin project \cite{pomerleau1989alvinn}.  In \cite{levine2016end}, a neural network is trained as a policy from images to manipulator arm torques using guided policy search.
Because these solutions do not separate image understanding and control, the learning dataset must provide examples of combinations of all control and visual inputs, requiring massive amounts of data for varying or complex control tasks.

Other solutions attempt to learn a drivability function directly from image data that can be used by a separate controller.  By utilizing accurate short range data provided by stereo vision, \citet{hadsell2009learning} learns a neural network to predict far-field traversibility from images, which is then fed into a separate planning and control framework.  However, this approach requires significant geometric image pre-processing. More recently, \citet{chen2015deepdriving} directly learns affordances necessary for autonomous driving by a low level controller.  However, these learned affordances cannot be used to drive with model predictive control framework such as \cite{williams2016} uses for aggressive control.

There has been a great deal of work into methods of semantic segmentation.  Lately, deep neural network architectures have achieved excellent results on semantic segmentation datasets such as \cite{arbelaez2012semantic} and \cite{long2015fully}.  These models aim to produce a per-pixel labeling of an input image.  Many techniques to improve the accuracy of these models, such as conditional random fields (CRFs) \cite{chen2014semantic} and dilated convolutions \cite{YuKoltun2016} have advanced the state of the art in this field.

\section{Approach}

Our approach combines a high performance control system based on Model Predictive Control (MPC), with deep Convolutional Neural Networks (CNNs) for real-time scene understanding.  We show that fully convolutional networks have the ability to go beyond the standard semantic image segmentation paradigm, and can generate a top-down view of the cost surface in front of the vehicle, even generalizing to portions of the track which are outside the camera's field of view, given a single video frame taken from the driver's perspective.

Model predictive control is an effective control approach for aggressive driving \cite{williams2017information, williams2016}. It is based on optimizing a cost function that defines where on a track surface the vehicle should drive. The cost surface must therefore encode the current and future positions of the road, obstacles, pedestrians, and other vehicles. This presents a major barrier for using MPC in novel environments since creating a cost function requires analyzing the local environment of the vehicle on-the-fly. Our solution is to train a deep neural network to transform visual inputs from a single monocular camera to a cost function in a local robot-centric coordinate system. In our implementation, the cost function takes the form of an occupancy-grid style cost map, as shown in Figure~\ref{fig:netArchitecture}. The network is trained so that the cost is lowest at the center of the track, and higher further from the center. This cost map can then be directly fed into a model predictive control algorithm.

In the categorization from \cite{chen2015deepdriving}, our approach is an example of direct perception.  This is in contrast to mediated perception, which uses geometry and 3D-reconstruction to project image-space labels into a planning space \cite{junior2008}, or the direct prediction of control outputs from input images as in behavior reflex (\cite{bojarski2016end}). Mediated perception requires the use of geometric transformations which can lose or distort  information.  Behavior reflex loses a great deal of generalization ability because the network must jointly encode vehicle dynamics and visual information, precluding the ability to generalize to different control tasks. 

By factoring the control and perception tasks, we can take advantage of the strengths and mitigate the weaknesses of both deep visual learning and MPC. The perception task of mapping images to cost functions is invariant to the control policy, which means that data can be collected from many different (off-policy) sources. This mitigates the main difficulty in deep learning, which is collecting large amounts of data. However, we are still able to use deep learning for on-the-fly scene understanding. In the case of model predictive control, we are able to operate without an explicitly programmed cost function, enabling its usage in potentially novel environments. However, we are still able to utilize MPC for the difficult problem of online optimization with non-linear dynamics and costs.

\subsection{Model Predictive Path Integral Control}

\begin{wrapfigure}{R}{0.3\textwidth}
\centering
\includegraphics[width=0.3\textwidth, trim = 0mm 167mm 50mm 17mm, clip = true]{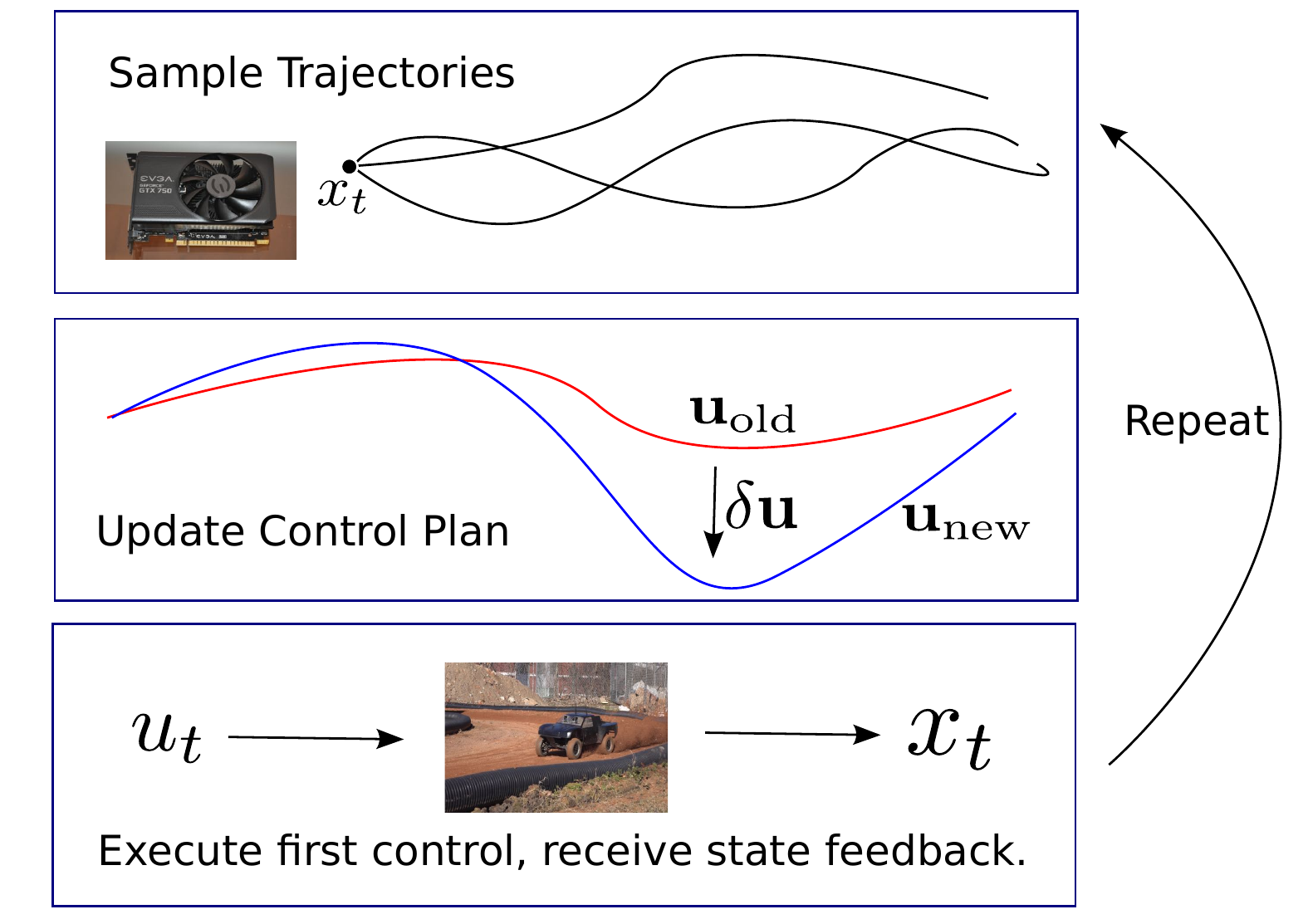}
\caption{Model Predictive Path Integral Control summary.}
\label{icml-historical}
\vskip -0.2in
\end{wrapfigure} 

Model predictive control works by interleaving optimization and execution: first an open loop control sequence is optimized, then the first control in that sequence is executed by the vehicle, and then state feedback is received and the whole process repeats. We use model predictive path integral control (MPPI), which is a sampling based, derivative free, approach to model predictive control which has been successfully applied to aggressive autonomous driving using learned non-linear dynamics \cite{williams2017information}.
At each time-step, MPPI samples thousands of trajectories from the system dynamics, and each one of these trajectories is evaluated according to its expected cost.  A planned control sequence is then updated using a cost-weighted average over the sampled trajectories.

Mathematically, let our current planned control sequence be $\left(u_0, u_1, \dots u_{T-1} \right) = U \in \mathbb{R}^{m \times T}$, and let $\left(\mathcal{E}_1, \mathcal{E}_2 \dots \mathcal{E}_K \right)$ be a set of random control sequences, with each $\mathcal{E}_k = \left(\epsilon_k^0, \dots \epsilon_k^{T-1}\right)$ and each $\epsilon_k^t \sim \mathcal{N}(u_t, \Sigma)$. Then the MPPI algorithm updates the control sequence as:
\begin{align}
\eta &= \sum_{k=1}^K \exp \left( -\frac{1}{\lambda} \left(S(\mathcal{E}_k) + \gamma \sum_{t=0}^{T-1} u_t^\rT \Sigma^{-1} \epsilon_k^t \right) \right) \\
U &= \frac{1}{\eta}\sum_{k=1}^K \left [ \exp \left( -\frac{1}{\lambda} \left(S(\mathcal{E}_k) + \gamma \sum_{t=0}^{T-1} u_t^\rT \Sigma^{-1} \epsilon_k^t \right) \right) \mathcal{E}_k \right]
\end{align}
The parameters $\lambda$ and $\gamma$ determine the selectiveness of the weighted average and the importance of the control cost respectively. The function $S(\mathcal{E})$ takes an input sequence and propagates it through the dynamics to find the resulting trajectory, and then computes the (state-dependent) cost of that trajectory sequence, which we denote as $C(x_0, x_1, \dots x_{T}) = \sum_{t=0}^T q(x_t)$. In this paper we only use an instantaneous running cost (there is no terminal cost), and we sample trajectories on a GPU using the dynamics model from \cite{williams2017information}. The instantaneous running cost is the following:
\begin{equation}
q(x) = w \cdot \left(C_M(p_x, p_y), (v_x - v_x^d)^2, 0.9^t I,  \left(\frac{v_y}{v_x}\right)^2 \right)
\label{equ:costFunction}
\end{equation}
where $C_M(p_x, p_y)$ is the output of the neural network which gives the track-cost associated with being at the body frame position $(p_x, p_y)$. The other terms are (1) A cost for achieving a desired speed $v_x^d$, (2) an indicator variable which is turned on if the track-cost, roll angle, or heading velocity become too high, and (3) is a penalty on the slip angle of the vehicle. The coefficient vector was $w = (100, 4.25, 10000, 1.75)$. Note that the three terms which are not learned are trivial to compute given the vehicle's state-estimate, while the cost map requires analysis of the vehicle's environment. In previous work \cite{williams2017information}, the cost map was obtained from a pre-defined map of the track combined with GPS localization, which does not generalize to other terrains.

\subsection{Convolutional Neural Network Architecture}
\label{sec:NetArchitecture}

\begin{figure}
\begin{center}
\includegraphics[width=0.4\columnwidth]{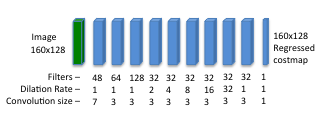}
\includegraphics[width=0.18\columnwidth]{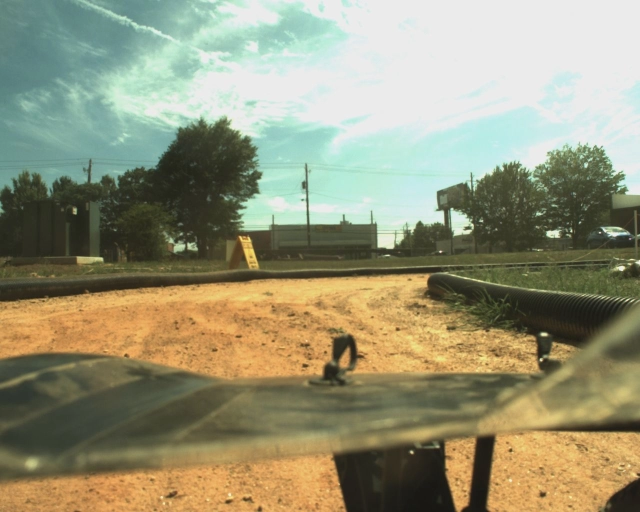}
\includegraphics[width=0.18\columnwidth]{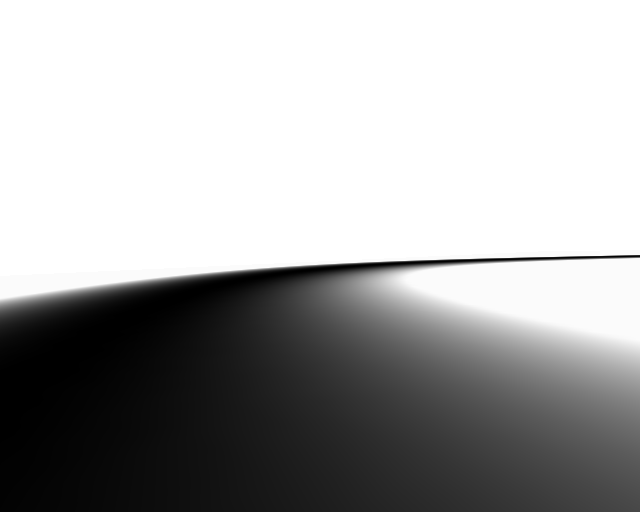}
\includegraphics[width=0.18\columnwidth]{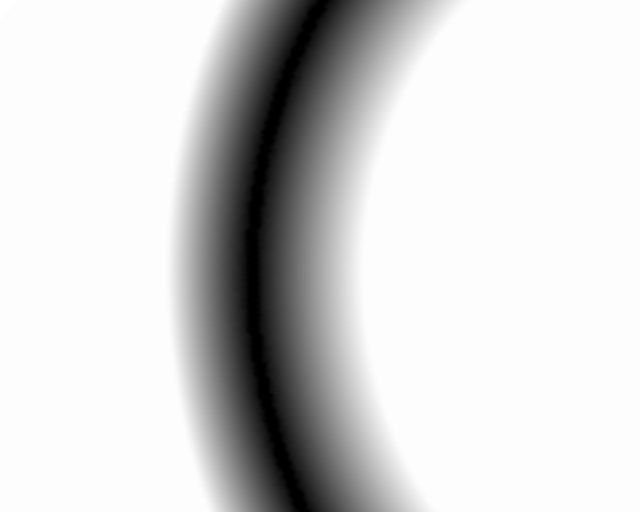}

\caption{Network architecture with input and training targets.  Left: Neural network architecture used to produce top down cost maps.  Right: example input image, image plane training target and top down training target, respectively}
\label{fig:netArchitecture}
\end{center}
\vskip -0.3in
\end{figure}


In this work, we use a CNN to generate costs based on future positions from a single monocular image.  Our CNN architecture is constrained to run in real time on the low power Nvidia GTX750Ti available on our platform, and it produces a dense costmap output. We found that a fully convolutional network that outputs a dense costmap with large input receptive fields produces the most accurate result. We trained this architecture to output two different types of predictions (as shown in Figure \ref{fig:overview}), these are (1) a top-down cost map that can be used directly by MPPI, and (2) an image-plane labeling of pixels that must be projected onto the ground before use.


We experimentally evaluate both the top down and image plane methods with two different neural network structures. The image plane network takes in 640x480 input images and passes them through several convolution layers and 2 pooling layers, followed by a set of 6 dilated convolution layers. The top down network uses a smaller structure, as shown in Figure \ref{fig:netArchitecture}. The dilated convolutions allow each output pixel the full input image as its receptive field while maintaining a reasonable (128x160) output size, this significantly improves the output quality of the network. The cost-map is then taken directly as the output of the final layer, with no normalization applied.  

Using these two network architectures, we are able to maintain low latency and a frame rate of about 10 Hz for the image plane network and 40Hz (full camera frame rate) for the top down network. Input images come directly from a PointGrey Flea3 color camera at 1280x1024 resolution.  These images are downsampled to 640x512, the dataset mean is  subtracted, and each pixel is divided by the dataset standard deviation.  During training, the 160x128 pixel output is compared with the pre-computed ground truth cost maps obtained from GPS data. It was found that the L1 loss produced a cleaner final costmap than L2 loss, so the training loss minimized is the sum of the L1 distances between the output pixel values and the ground truth pixel values.  This loss is only computed for points within 10 pixels of the edge of the track in the ground truth image, this is done to avoid training the network to output large sections of blank space.  The network was trained using the Adam \cite{kingma14adam} optimization algorithm in Tensorflow \cite{tensorflow2015}.  A mini-batch size of 10 images was used during training, and a small random perturbation to the white balance of each image (multiplying each channel by a normally distributed random variable between 0.9 and 1.1) was also applied. For all networks, best driving performance was achieved with training stopped at or near 100,000 iterations. This coincided with the point where testing loss on a held out dataset plateaued.

\subsection{Ground truth generation} \label{gtGeneration}

\begin{figure}
\vskip 0.2in
\begin{center}
\includegraphics[width=0.356\columnwidth]{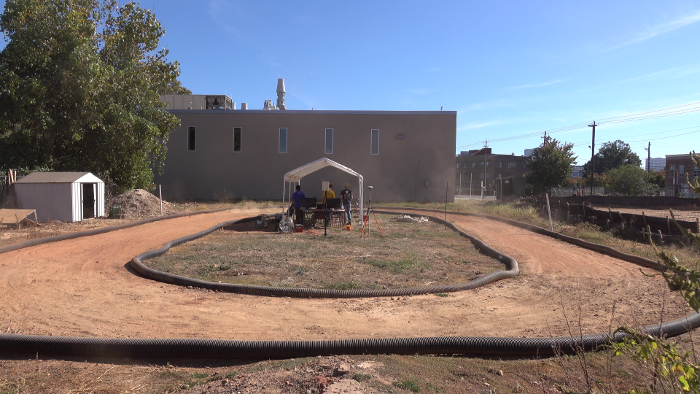}
\includegraphics[width=0.3\columnwidth]{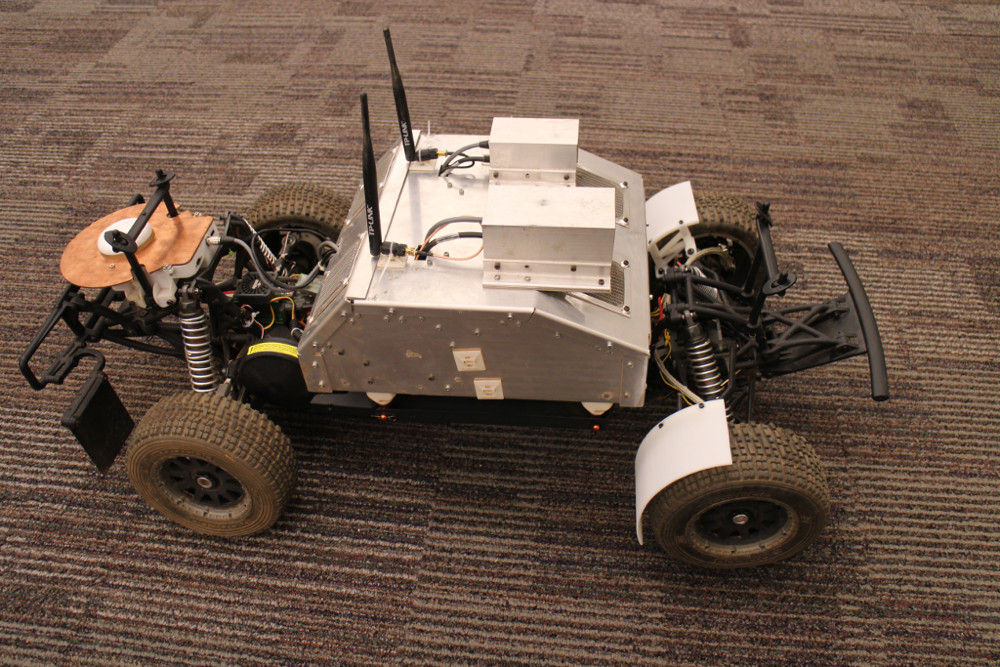}
\includegraphics[width=0.3\columnwidth]{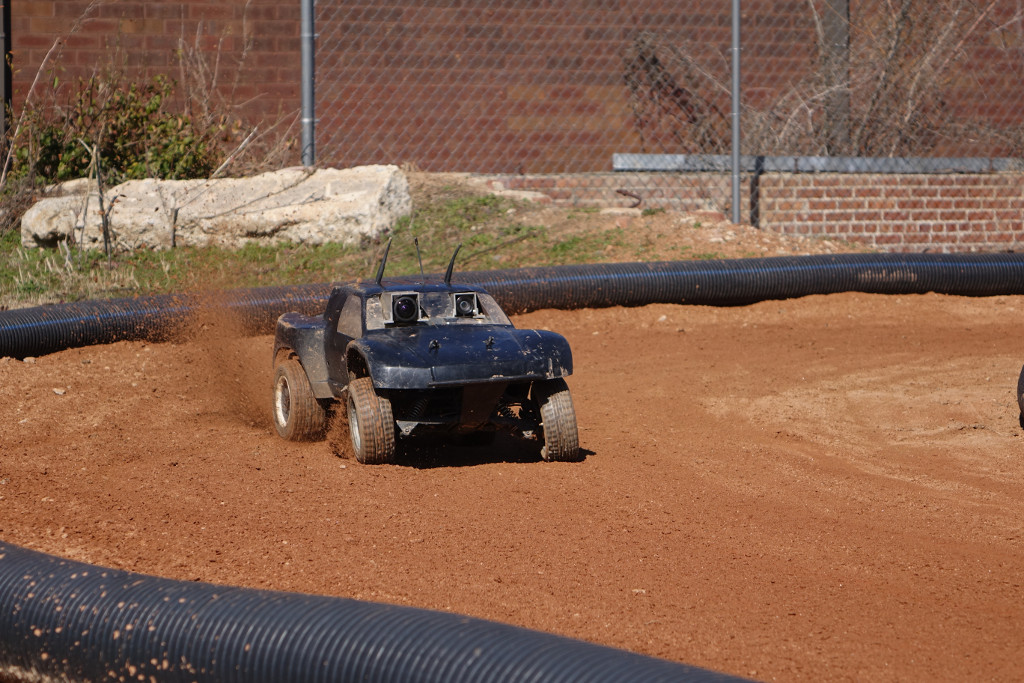}
\caption{Testing setup.  Left: Oval dirt test track where all test data was taken.  Center: Experimental vehicle, with cameras, onboard computation, and GPS/IMU position system.  Right: Photo of vehicle during testing.}
\label{fig:testingsetup}
\end{center}
\vskip -0.2in
\end{figure} 

\begin{figure}
\begin{center}
\includegraphics[width=0.17\columnwidth]{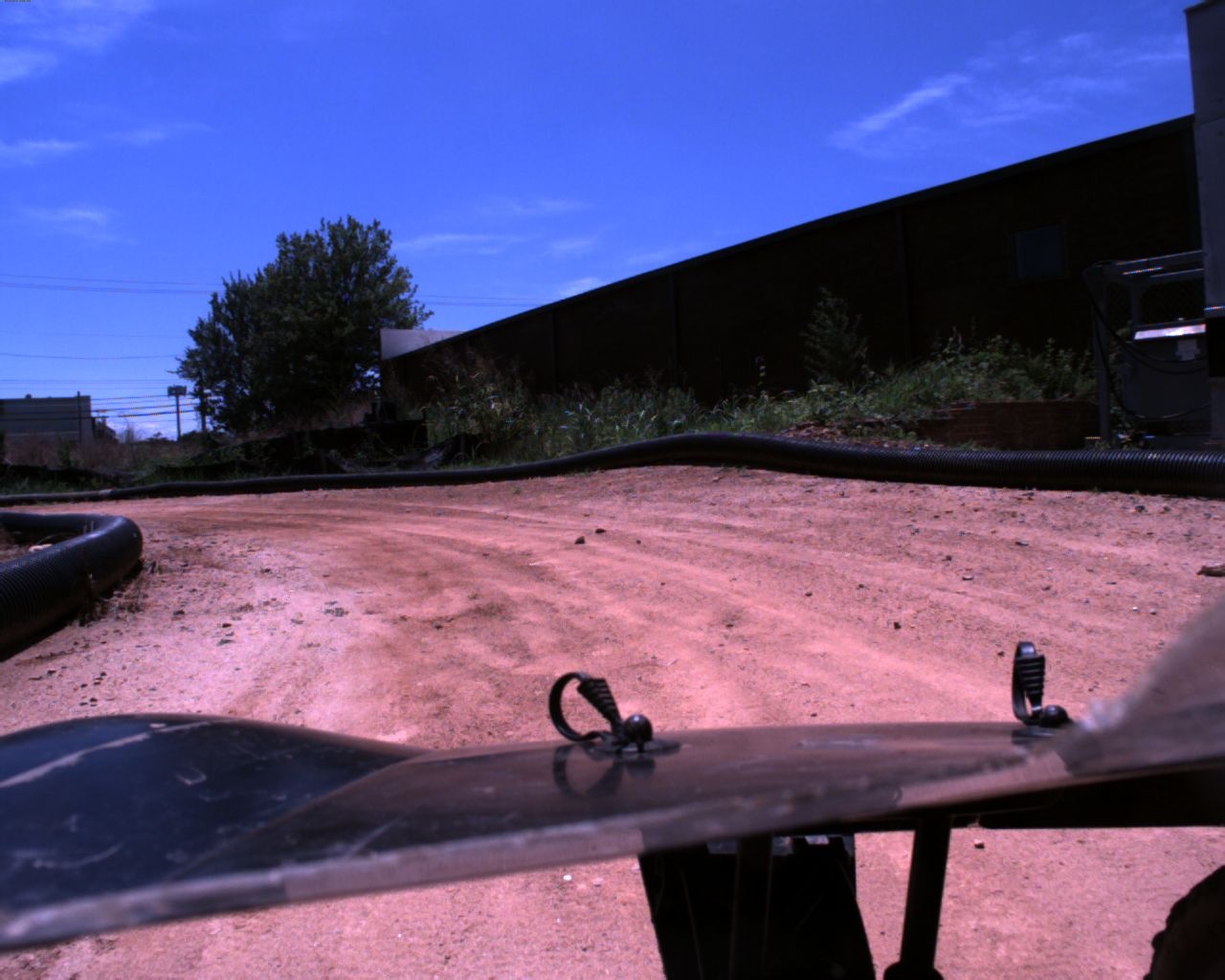}
\includegraphics[width=0.17\columnwidth]{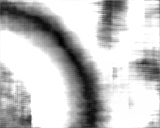}
\includegraphics[width=0.17\columnwidth]{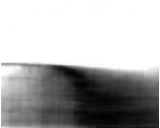}
\includegraphics[width=0.17\columnwidth]{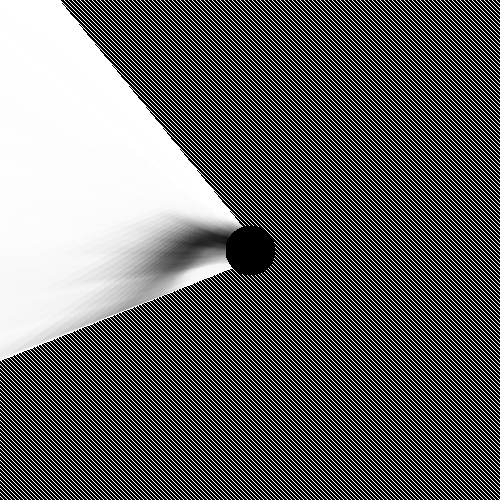}
\caption{Neural Network output examples. Top: Input RGB image and top-down costmap Bottom: Image Plane neural network output and reprojection to ground plane (White cone is camera field of view).}
\label{fig:reproject}
\end{center}
\end{figure}


In order to learn a pixel-wise regression function capable of producing traversal costs at every pixel, training data is needed on the order of 100s of thousands of frames.  Labeling all of this data by hand is laborious, slow, and prone to errors.  However, the 1:5 scale AutoRally vehicle (Fig. \ref{fig:testingsetup}) that we use in our experiments, and many autonomous and commercial vehicles are equipped with position sensors and cameras that can associate each image with a full state estimate, including orientation and position. Combined with a surveyed map of a track registered to GPS coordinates, these can be used to create hundreds of thousands of labeled images without any manual labeling of individual images.

By calibrating the transformation between the IMU (where position and orientation estimates are calculated) and the camera, a homography matrix can be computed that transforms the surveyed track map from world coordinates to image plane coordinates:

\begin{equation}
H = k T^{car}_{im} T^{world}_{car}
\end{equation}

Where $T^{world}_{car}$ is the position of the car in world coordinates (estimated at the IMU), $T^{car}_{im}$ is the transformation between the IMU and camera reference frames, and $k$ is the camera intrinsics matrix.  Given this, points in the ground coordinate frame can be projected into the image using:

\begin{equation}
p_{im} = \hat{H} p_{world}; \hat{H} = 
\begin{bmatrix}
    	H_{11}       & H_{12} & H_{14} \\
    	H_{21}       & H_{22} & H_{24} \\
    	H_{31}       & H_{32} & H_{34}
\end{bmatrix}
\end{equation}

Where $p_{im}$ and $p_{world}$ are homogenous points.  
Using this scheme, ground truth images can be produced for each image in our training set.  This mapping in not perfect due to small errors in time synchronization and violations of the assumption of constant height above ground of the camera.  However, despite these small errors, the reprojected cost maps are very good, and networks are able to learn from them.  To produce ground truth images for the top down network, a 160x128 section of the cost map directly in front of the vehicle(in vehicle centric coordinates) is used.

Using this method, we created approximately 300,000 images with corresponding ground truth cost maps.  These training images were taken from 64 different runs spanning 9 different days over the course of 8 months.  It includes substantial variability in lighting conditions, people and equipment present at the collection site, and poses of the camera on the track.  This data is split into approximately 250,000 training images and 50,000 test images.

\subsection{Implementation}

In order to truly test the performance of a neural network designed for autonomous driving, it must be implemented and tested on a physical platform.  While testing on datasets can show a great deal about how a neural network might perform and in what cases it may fail, there is no substitute for real-world testing on a robot.  In our case, we choose the AutoRally platform (see Figure \ref{fig:testingsetup}).  This is based on a 1:5 scale RC chassis capable of aggressive maneuvers and a top speed of nearly 60 miles per hour.  It includes all sensors and computation required onboard the vehicle.  During testing, all computation is performed on the vehicle in real time, including neural network forward inference and model predictive control computation.  The primary functions of the testing setup are:
\begin{itemize}
        \item Image capture using onboard Point Grey cameras
	\item Deep model forward inference in Tensorflow to produce a cost map from an image
	\item Generation of homography using IMU based orientation and velocity
	\item Trajectory cost lookup and control computation
	\item Forward propagation of controller state, so a single image can continue to be used by the controller until a new image is available from the deep model.
\end{itemize}

All software is written using the Robot Operating System (ROS) framework and custom software packages to communicate with the AutoRally system.  Forward inference through the network is handled asynchronously, and the cost maps are fed to the MPPI control algorithm at approximately 10Hz for the image plane network and 40Hz for the top down network, and the MPPI controller runs at 40Hz.  Velocity and acceleration information is obtained from the onboard GPS-IMU system, but absolute position is not used.  For the top down case, the camera orientation (from the IMU) is used to generate a homography matrix that will transform ground plane points in trajectories generated by MPPI into image plane points that can be be assigned a positional cost based on the output of the deep neural network.  The neural network output and associated homography matrix are used by the MPPI algorithm to plan and execute controls until another cost image is available.  All of the computation for MPPI and the CNN happen on the same GPU on-board the robot.


\section{Experimental Results}

In order to evaluate the performance of the proposed system, we tasked the CNN-MPPI algorithm with driving around a roughly elliptical dirt track, using 1/5 scale vehicle hardware similar to \cite{williams2016, williams2017information}. This enables us to compare lap times and speeds achieved with the same controller using a ground truth cost-map.  This provides a metric, independent of network validation error, of how well the neural network performs in a real world scenario. We compare both of our costmap generation methods, top down (TD) and image plane (IP), in this testing scenario. Additionally, we compare the performance of the convolutional neural networks (mean L1 pixel distance)on a held out testing data set to gain some insight into performance discrepancies and failure modes.

\subsection{Network Performance}

The accuracy of the neural networks is computed as the L1 distance between the ground truth and the training target on a holdout dataset of approximately 4000 images. In order to achieve a meaningful metric, we report only the error for pixels where we there is track (i.e. anywhere the ground truth training image is not white).  We use this convention in all neural network training we report.  We report this as: $\text{score} = (1 - \text{error})$.

The top down network, which maps input images to a top down (bird's eye) costmap achieved a score of 0.92. The image plane network, which maps input images to an \emph{image plane} costmap achieved a score of 0.82. In addition the having a lower score than the top down network, the image plane costmap must also be projected onto the ground plane in order to be usable as a cost function. This projection operation adds to the overall complexity of the method and introduces additional errors, the conclusion is therefore that the top down network is both more accurate and easier to implement than the image plane approach. Note that both the top down and image plane networks used the exact same input training set. 

\begin{figure}
\begin{minipage}{0.5\columnwidth}
\includegraphics[width=0.95\columnwidth]{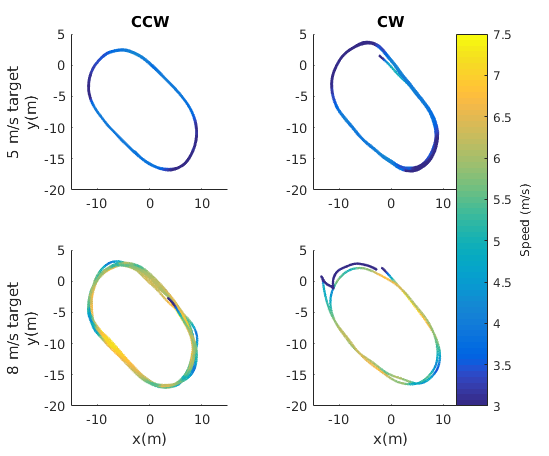}
\end{minipage}
\begin{minipage}{0.5\columnwidth}
\begin{tabular}[c]{cc}
\includegraphics[width=0.48\columnwidth]{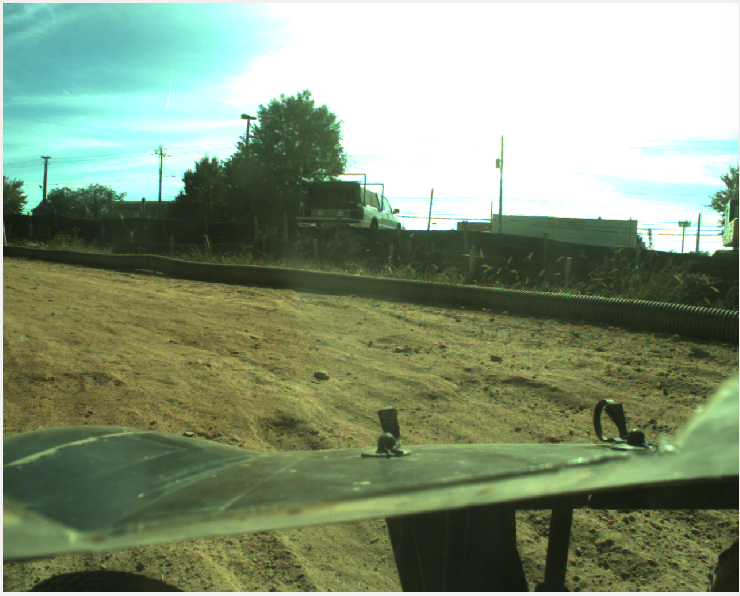}
\includegraphics[width=0.48\columnwidth]{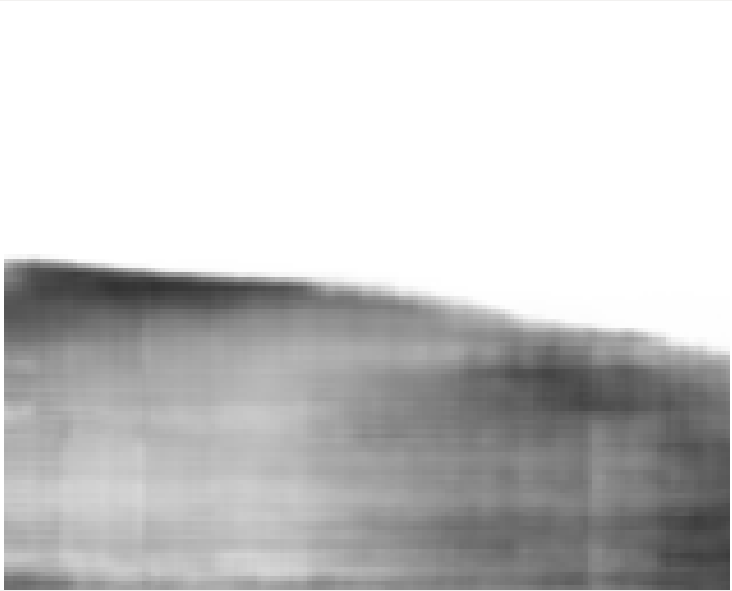} \\
\includegraphics[width=0.48\columnwidth]{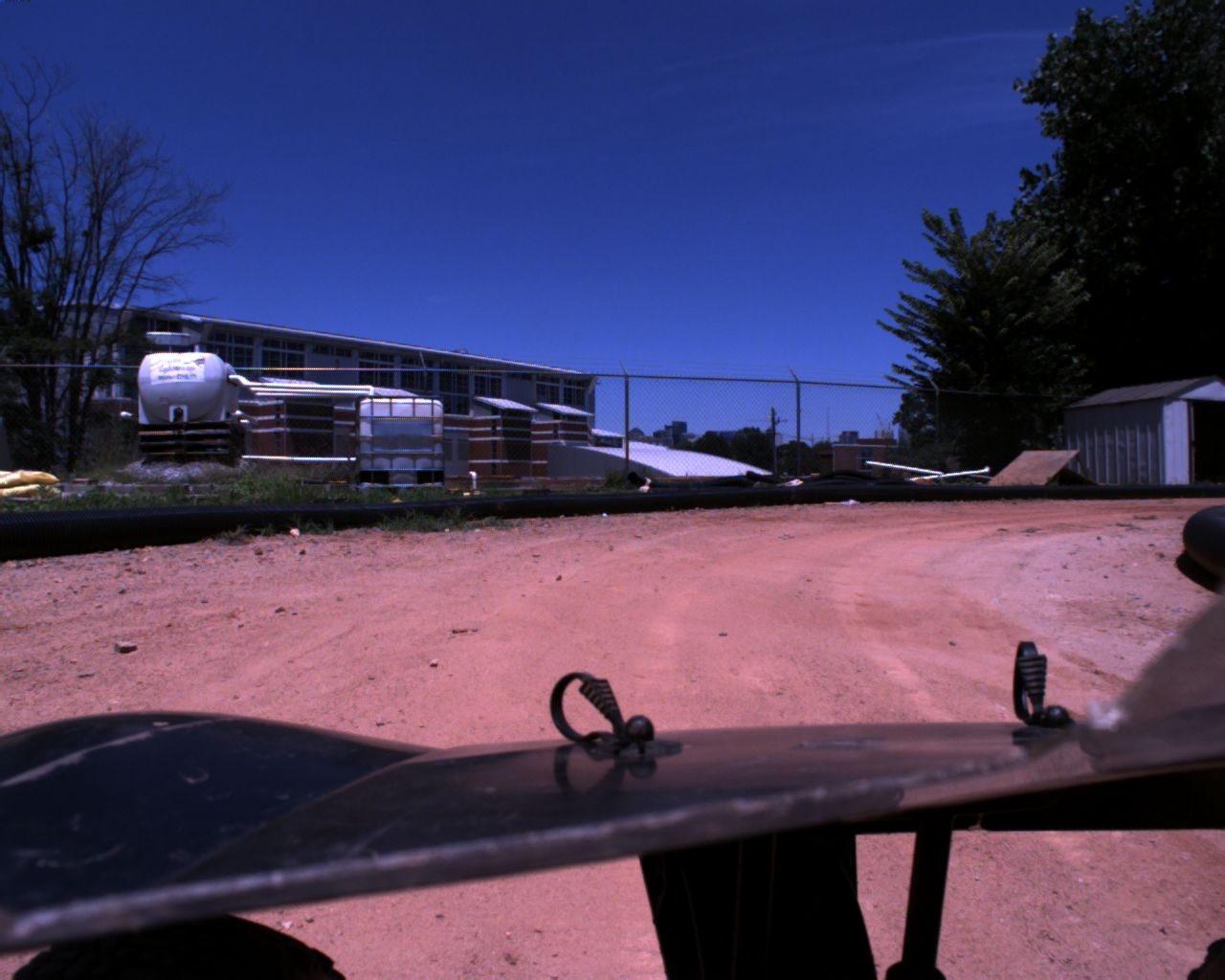}
\includegraphics[width=0.48\columnwidth]{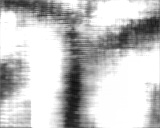}
\end{tabular}
\end{minipage}
\caption{Left: GPS plots of vehicle trajectory with top-down network at 5 m/s and 8 m/s target speeds.  Notice how the method is able to reject strong disturbances at the limits of vehicle handling.  Right: Failure mode, where the network sees track that is not there.}
\label{fig:velocities}
\vskip -0.2in
\end{figure}

Figure \ref{fig:velocities} illustrates a network prediction failure resulting in a vehicle crash for both the top down network and the image plane network.  The largest takeway from this is that the top-down network seems to still produce plausible images even in failure, whereas the results of the image plane network are unusable for planning.  

\subsection{Ablation Study}

\begin{figure}
\vskip 0.2in
\begin{center}
\includegraphics[width=0.2\columnwidth]{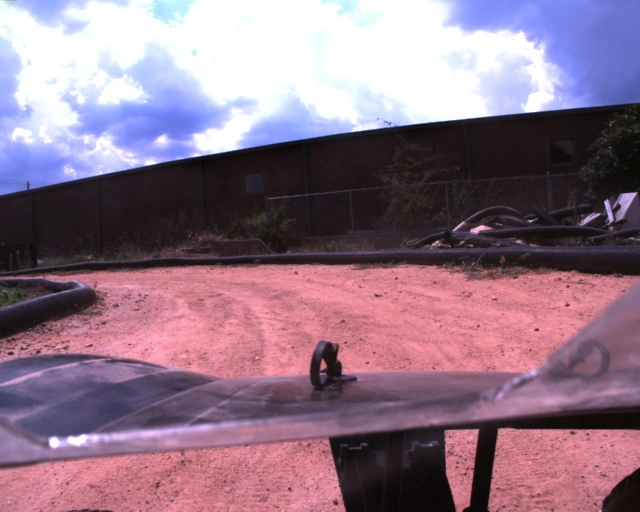}
\includegraphics[width=0.2\columnwidth]{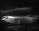}
\includegraphics[width=0.2\columnwidth]{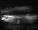}
\includegraphics[width=0.2\columnwidth]{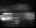}
\includegraphics[width=0.2\columnwidth]{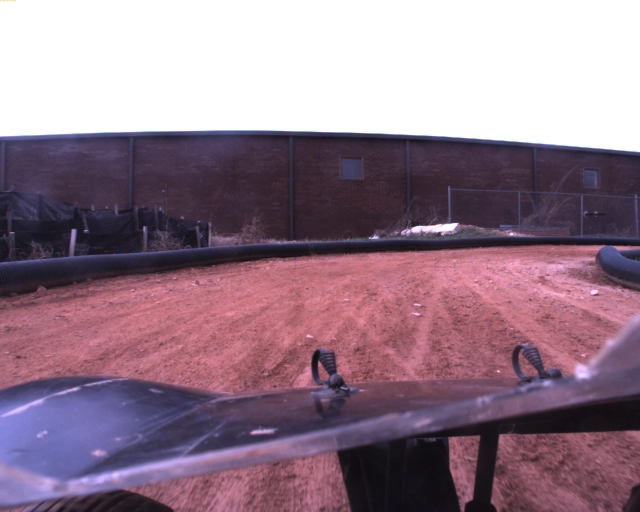}
\includegraphics[width=0.2\columnwidth]{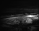}
\includegraphics[width=0.2\columnwidth]{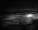}
\includegraphics[width=0.2\columnwidth]{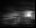}
\caption{Ablation Study.  Examples of (left to right) input image, ablation histogram for box size of 6, 10, and 20 pizels.  It is clear in both cases that the inside of the corner is the most important feature for determining track shape, with the top of the building providing some clues as well. }
\label{fig:ablation}
\end{center}
\vskip -0.2in
\end{figure} 

We performed an ablation study in order to identify the features of the input images that play the strongest role in the generation of the cost map. We first obtain as a baseline the costmap which is generated by the network from the full input image.  We then "zero out" a block of pixels at a certain location and size by replacing all pixels within the window with the mean pixel value from the entire dataset. After mean subtraction, this block will have the value zero and will therefore not contribute to the network activations. We systematically examine the influence of different parts of the image on the prediction performance by scanning the window over the entire input image, thereby generating a set of ablation images with zeroed out blocks at different locations. For each ablated image, we compute an accuracy measure (average L1 distance for track pixels). We then construct a sensitivity map by creating an image from these accuracy measures, which each accuracy value is stored at the corresponding location where a block was ablated. Figure \ref{fig:ablation} shows several sensitivity maps for representative input images and ablation block sizes.  All images are normalized with zero error as black and largest recorded error as white.

This study demonstrates that the network has learned to use intuitively reasonable input features such as the left and right track boundaries. The network can tolerate the removal of small track regions due to  ablation and still produce usable cost maps. It can also produce accurate results when unimportant image regions, such as sky, are removed.

\subsection{Driving Performance}

\begin{table}
\centering
\caption{Testing statistics for image plane (ip) and top down (td) networks}
\label{tab:speed_results}
\small
\begin{tabular}{c|c|c|c|c|c|c}
\hline
Method& \multicolumn{2}{|c|}{Counterclockwise travel} & \multicolumn{2}{|c|}{Clockwise travel} \\ \hline
& Avg. Lap (s))& Top Speed (m/s) & Avg. Lap (s)& Top Speed (m/s)\\ \hline \hline
(TD) 5 m/s& 16.98       & 4.37              & 18.09    & 4.99  \\ \hline
(TD) 6 m/s& 12.19        & 6.38               & failure     & failure  \\ \hline
(TD) 7 m/s& 10.84        & 6.91              & 11.27     & 6.51  \\ \hline
(TD) 8 m/s& 10.13        & 7.47               & failure    & failure  \\ \hline
(IP) 6 m/s& 14.48 & 5.67 & failure & failure \\ \hline
\cite{williams2017information}& 9.74 & 8.44 & N/A & N/A \\ \hline
\cite{williams2016} & N/A & N/A & 10.04  & 7.98 \\
\end{tabular}
\end{table}


The goal in learning to predict a costmap is to be able to use them to maneuver the vehicle effectively. In order to test the overall effectiveness of our approach, we conducted an experiment in which we attempted to drive a 1/5 scale AutoRally vehicle at increasingly aggressive speeds around a flat dirt track. Each method uses the same controller and vehicle physics model, cameras, and track. The form of the controller's cost remains the same, although some parameters are tuned slightly to optimize performance for each method.  To find the limits of each method, we slowly increased the target speed from 5m/s, performing 10 laps at each target speed, until the method was no longer able to remain on the track. Note that the friction limits of the vehicle going around the tracks turns are around 5.5 m/s, so the control algorithm has to intelligently moderate both the steering and throttle in order to navigate successfully.  

Using the top down network produced significantly more robust, consistent, and overall faster runs than the image plane network. Using the image plane network, the it was only occasionally possible to produce runs of 10 consecutive laps (at the slowest speed). Most of the runs lasted between 1 and 5 laps before intervention was required. Usually, this was due to the network not identifying a turn, which would result in the vehicle driving to the end of the track and stopping. Figure \ref{fig:reproject} demonstrates the difference between the two approaches as the vehicle approaches a turn, the top-down network produced much cleaner and crisper costmaps in corners where only a small portion of the track is visible.  

Table \ref{tab:speed_results} summarizes lap times and top speeds for our networks and the the method in \cite{williams2016, williams2017information}. In \cite{williams2017information}, using GPS localization in a pre-defined map, the vehicle and controller were able to achieve an average lap time of 9.74 seconds, only 0.39 seconds faster than the best setting of our method which only uses a single monocular image, body frame velocity, and inertial data as input. The image plane regression network was able to achieve a maximum average lap time of 14.48 seconds over 10 laps, 4.74 seconds slower. This difference is due to the top-down network producing crisper output costmaps, as well as its ability to predict beyond the field of view.

\section{Conclusions}

In this work, we present field experiments demonstrating novel capabilities of fully convolutional neural networks combined with sampling based model predictive control.  We compare two output targets for the neural network, a cost map projected into the image plane and a top down view of the cost map, and find that the top down network only loses 17\% lap time over ground truth, while the image plane network loses 29\%.  We compare them both on a sample of a held-out dataset and in full system experiments driving an autonomous vehicle.

Surprisingly, the fully convolutional network is able to learn the complex, out of plane transformation needed to project the observed image pixels to the ground plane and predict a map of drivable area in front of the vehicle.  Based on previous research, we initially assumed that CNN's would be better suited to predicting costs in the image plane, where the position invariant nature of CNNs can be fully utilized.  However, the network was able to extremely accurately reproduce this out of plane representation, which often included information not directly visible in the image.  For example, it is able to produce track estimates further around corners than the camera field of view can directly observe.

This ability to predict around corners was critical in performance for the controller.  The model predictive controller only uses information that it can see in the output of the neural network, and plans ahead 1.5 seconds to produce a control signal.  This 1.5 second time horizon leads to extremely timid behavior in the case of the image plane regression network because the available look ahead distance is very short.  This was not the case with the network that directly regressed the top-down view, and was a large contribution to its success in vehicle performance.

Additionally, the top-down network tends to produce a map with a defined centerline that is still good for planning, even if the exact location of the track is incorrect.  This allows the MPPI algorithm to continue planning feasible paths until another image is processed, hopefully rectifying the errors.

\clearpage


\bibliography{path_integral_bib,deepnet}

\end{document}